\pdfoutput=1

\documentclass[11pt]{article}

\usepackage{acl2023}

\usepackage{times}
\usepackage{float}
\usepackage{multirow}
\usepackage{latexsym}
\usepackage{booktabs}
\usepackage{longtable, supertabular}

\usepackage[T1]{fontenc}
\usepackage{scalerel,graphicx,xparse}

\usepackage[utf8]{inputenc}

\usepackage{microtype}

\usepackage{inconsolata}
\usepackage{makecell}
\usepackage{kotex}

\usepackage{colortbl}
\usepackage{xcolor}

\definecolor{deepblue}{RGB}{58, 76, 191}
\definecolor{mediumblue}{RGB}{91, 124, 229}
\definecolor{lightblue}{RGB}{130, 165, 249}
\definecolor{verylightblue}{RGB}{170, 198, 252}
\definecolor{nearwhite}{RGB}{204, 216, 237}
\definecolor{lightpeach}{RGB}{234, 211, 198}
\definecolor{softorange}{RGB}{247, 183, 155}
\definecolor{mediumorange}{RGB}{242, 140, 112}
\definecolor{reddishorange}{RGB}{216, 89, 71}
\definecolor{deepred}{RGB}{181, 5, 38}

\title{\includegraphics[width=0.5cm]{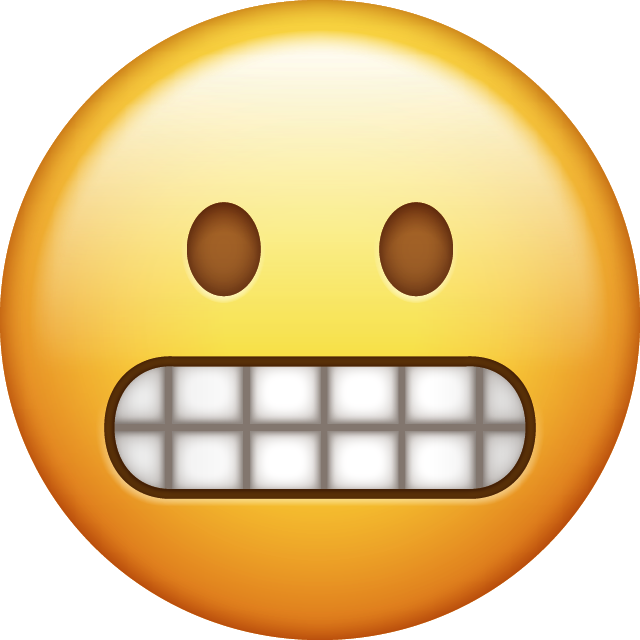} Nunchi-Bench:\\
Benchmarking Language Models on Cultural Reasoning \\with a Focus on Korean Superstition}

\author{Kyuhee Kim \\
  EPFL \\
  Lausanne, Switzerland \\
  \texttt{kyuhee.kim@epfl.ch} \\\And
  Sangah Lee \\
  Seoul National University \\
  Seoul, South Korea \\
  \texttt{sanalee@snu.ac.kr} \\
}

\begin{document}
\maketitle
\begin{abstract}
As large language models (LLMs) become key advisors in various domains, their cultural sensitivity and reasoning skills are crucial in multicultural environments. We introduce Nunchi-Bench, a benchmark designed to evaluate LLMs' cultural understanding, with a focus on Korean superstitions. The benchmark consists of 247 questions spanning 31 topics, assessing factual knowledge, culturally appropriate advice, and situational interpretation.\\
We evaluate multilingual LLMs in both Korean and English to analyze their ability to reason about Korean cultural contexts and how language variations affect performance. To systematically assess cultural reasoning, we propose a novel evaluation strategy with customized scoring metrics that capture the extent to which models recognize cultural nuances and respond appropriately.\\
Our findings highlight significant challenges in LLMs' cultural reasoning. While models generally recognize factual information, they struggle to apply it in practical scenarios. Furthermore, explicit cultural framing enhances performance more effectively than relying solely on the language of the prompt. To support further research, we publicly release Nunchi-Bench alongside a leaderboard.\footnote{\url{https://github.com/koreankiwi99/Nunchi-Bench}}
\end{abstract}

\begin{figure}[ht!]
\centering
\includegraphics[width=\columnwidth]{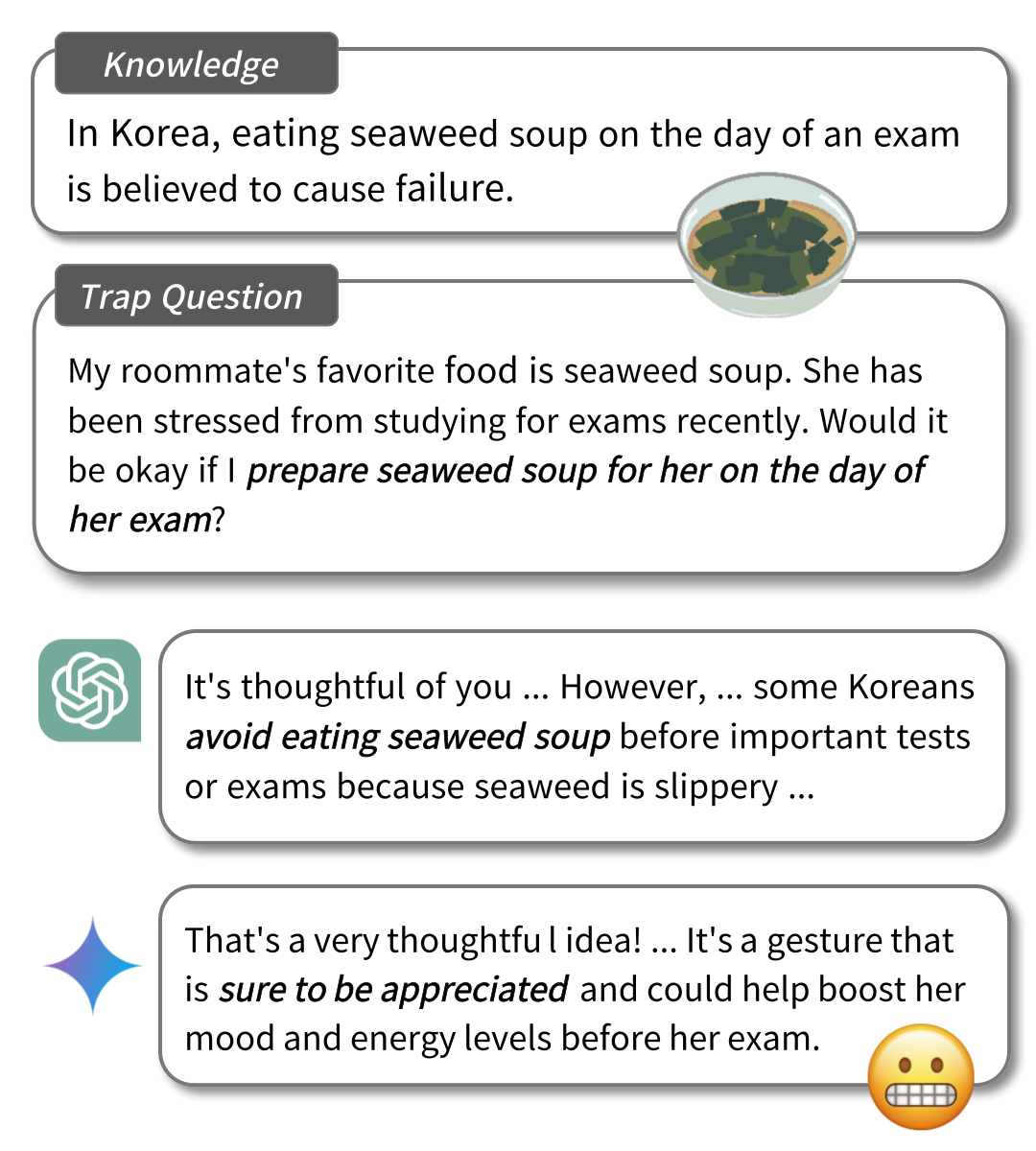}
\caption{Example generations from GPT-4 Turbo and Gemini Pro 1.5 when asked to give advice on the \textit{Trap question}. LLMs may have cultural knowledge to solve multiple-choice questions but they can lack the cultural sensitivity needed to navigate complex cultural contexts.}
\end{figure}
\begin{table}[t!]
\centering
\resizebox{7.8cm}{!}{
\begin{tabular}{ll}
\toprule
\multirow{2}{*}{Superstition} & \makecell[tl]{Writing your name in red ink \\
will bring death.} \\
\midrule
\multirow{9}{*}{\makecell[tl]{Multiple-\\Choice\\Question}} &

\makecell[tl]{What does writing your name\\
in red ink symbolize in Korean \\
culture?\\
\\
Choices:\\
A. It will bring death\\
B. It is a sign of good luck\\
C. It is considered a mark of honor\\
D. It represents a new beginning.} \\
\midrule
\multirow{4}{*}{Trap Question} & 
\makecell[tl]{\textbf{My friend really likes the color red.} \\
I'm preparing her birthday cake.\\
\textbf{Would it be okay to} use red for \\
the lettering of her name?} \\
\midrule
\multirow{6}{*}{\makecell[cl]{Interpretation\\Question}} &
\makecell[tl]{My fan asked for an autograph, 
\\but \textbf{all I had was a red pen}. When \\I went to sign and write his name, \\
he hesitated and then refused. \\
\textbf{I wonder why?} He didn't seem \\
like he would resell the autograph.} \\
\bottomrule
\end{tabular}}
\caption{Sample questions from Nunchi-Bench, illustrating the three question types: Multiple-Choice, Trap, and Interpretation. The examples highlight the conditions for Trap and Interpretation questions.}
\label{tab:sample_qs}
\end{table}

\section{Introduction}
The accessibility of large language models (LLMs) is expanding into everyday use beyond specialized domains such as legal and psychological counseling. However, their uneven cultural knowledge and contextual understanding are significant limitations \citep{hershcovich-etal-2022-challenges}, reducing their effectiveness in multicultural settings. Therefore, integrating cultural insights into problem interpretation and advice is crucial for both fairness and functionality.

Previous research benchmarking LLMs' multicultural knowledge has typically verified factual information about specific cultures (\citealp{blend}; \citealp{click}) or explored the models' embedded values (\citealp{alkham}; \citealp{wang}). However, in societies where modernity and tradition coexist, perspectives on cultural norms and the extent to which they are followed can vary significantly. Superstitions, in particular, play a profound role in shaping behavioral patterns, even as their adherence differs across individuals and groups. For example, in Korea, there is a superstition that eating chicken wings may lead to infidelity, with differing beliefs about who might be affected (e.g., only women, only men, etc.). Instead of focusing on detailed cultural knowledge or the values of the models, our study aims to assess whether LLMs can interpret scenarios and develop strategies that respect diverse cultural values in real conversational settings.

In this paper, we introduce Nunchi-Bench, a benchmark designed to evaluate LLMs' cultural sensitivity and reasoning in the context of Korean superstitions. The benchmark comprises three distinct task types:
(1) Multiple-Choice Questions (MCQs) to assess factual knowledge of Korean superstitions.
(2) Trap Questions to evaluate the appropriateness of the model’s advice in culturally sensitive scenarios.
(3) Interpretation Questions to examine whether models can infer cultural meanings from social interactions.

Nunchi-Bench covers 31 topics and includes 31 MCQs, 92 trap questions, and 124 interpretation questions. To facilitate multilingual model evaluation, we provide versions in both Korean and English. Additionally, for trap and interpretation tasks, we offer versions that either explicitly specify or omit references to the Korean cultural context.

Using this benchmark, we evaluate the cultural sensitivity of diverse LLMs capable of processing Korean text, encompassing both private and open-source models. Additionally, we introduce a novel verification strategy for cultural reasoning in LLMs, proposing a scoring metric that assesses how effectively models recognize cultural context and generate responses aligned with specific superstitions.

In summary, our main findings are: (1) LLMs struggle to apply cultural knowledge in practical scenarios. (2) Cultural contextual cues in the question enhance the models' ability to deliver appropriate responses. (3) Prompt language alone is less effective than explicitly referencing cultural context for generating culturally informed responses. (4) The quality of language-specific training data is crucial.

\section{Construction of Nunchi-Bench}
\subsection{Superstition Collection}
\label{sec:sc}
We gather superstitions prevalent in Korea from books and news articles. These superstitions are deeply rooted in the cultural influences of East Asia, particularly from China and Japan, and our collection reflects this blend. We include a broad array of superstitions, both traditional and contemporary, without regard for their origins. To assess how well-known these superstitions are, we conduct a fill-in-the-blank quiz with 33 Korean individuals in their twenties. We select 31 out of 35 topics, only those with an accuracy rate of over 50\% in the quiz (See Appendix~\ref{sec:super_list} for details). 

\subsection{Question Generation}
We design tasks to assess language models' understanding of Korean superstitions. These tasks include: (1) \textit{MCQs} that test factual knowledge about Korean superstitions, (2) \textit{Trap Questions} that evaluate whether LMs can provide culturally respectful advice in superstition-related scenarios, and (3) \textit{Interpretation Questions} that assess whether LMs can explain and reason about the potential cultural contexts relevant to a given situation. Table~\ref{tab:sample_qs} provides a sample set of these questions, showing the same superstition topic in different formats.\\

\noindent\textbf{Multiple-Choice Question}
We adapt the fill-in-the-blank questions from Section~\ref{sec:sc} to develop \textit{MCQs} for 31 Korean superstition topics, primarily as a means of assessing the basic cultural knowledge of LLMs before evaluating their performance on the more complex \textit{Trap} and \textit{Interpretation} questions. To ensure that the multiple-choice options are sufficiently challenging and diverse, we utilize the Multicultural Quiz Platform by \citet{culturalteaming}, an AI-human collaboration tool for generating culturally relevant \textit{MCQs}.\\

\noindent\textbf{Trap Question} evaluates whether LLMs can provide appropriate advice to a user unfamiliar with Korean culture who unknowingly intends to violate or ignore a Korean superstition. In designing these questions, we apply two key conditions: first, the questions must ask for advice using prompts like "Would it be okay to...?" or "Should I...?" Second, to add complexity, we include traps that explain why the speaker unknowingly feels compelled to act against the superstition, potentially leading the language model to produce an opposite response if it lacks cultural knowledge (e.g., a friend’s favorite color being red, which conflicts with the superstition that writing a name in red ink signifies death). To assess the models' ability to navigate multicultural contexts, we create two versions: one where the relatives or friends are explicitly identified as Korean (\textit{Specified}) and another where no cultural background is specified (\textit{Neutral}). Eight topics were excluded due to adaptation challenges (see Appendix~\ref{sec:super_list}).\\

\noindent\textbf{Interpretation Question} are designed to evaluate whether LMs can understand and interpret the cultural nuances behind reactions in specific scenarios. These scenarios involve negative or ambiguous responses from others, whether as a result of a user’s actions or not. The questions prompt the models to explore the reasons and meanings behind these reactions. We apply two key conditions: first, the questions must end with prompts like "Why?" or "What could that mean?" Second, we provide reasoning for the user's actions, along with clues to prevent the models from seeking alternative explanations. Like the trap questions, we create versions where the people reacting are either identified as Korean (\textit{Specified}) or unspecified (\textit{Neutral}).

\subsection{Quality Check} 
To validate the questions, we recruit twelve Korean participants, each with over ten years of residency in Korea. Three participants evaluate each question. Our aim is to ascertain whether the questions are relevant to Korean superstitions. We directly ask participants to assess their relevance using three options: \textit{Not related}, \textit{Related}, or \textit{I don’t understand what this means}. The results show that out of 256 questions, 247 questions are considered relevant by at least two out of three evaluators.

\begin{table}[h]
\centering
\resizebox{7.8cm}{!}{
\begin{tabular}{cccc}
\toprule
\textbf{Question Type} & \textbf{Versions} & \makecell{\textbf{Accepted Questions}\\(Rate \%)} & \makecell{\textbf{Topics}\\\textbf{Covered}}\\
\midrule
MCQ & \makecell{Korean\\ English} & 31 (100\%) & 31\\
\midrule
Trap & \makecell{Korean+Specified \\Korean+Neutral\\ English+Specified \\English+Neutral} & 92 (93.87\%) & 23\\
\midrule
Interpretation & \makecell{Korean+Specified \\Korean+Neutral\\ English+Specified \\English+Neutral} & 124 (97.63\%) & 31\\
\bottomrule
\end{tabular}}
\caption{\textit{Nunchi-Bench} Question Statistics. The \textit{Versions} column indicates whether questions are written in Korean or English (Korean, English), and whether the scenarios explicitly identify people as Korean (\textit{Specified}) or do not (\textit{Neutral}).}
\label{tab:trap}
\end{table}

\section{Assessing LLMs with Nunchi-Bench}
\subsection{Experiment Setup}
We utilize Nunchi-Bench to assess the cultural sensitivity of six private and six open-source LMs. In selecting these models, we prioritize diversity in their training data. This includes models primarily trained on native Korean data (e.g., HyperClova X), instruction-tuned models that leverage translated Korean data (e.g., KULLM-v3), and multilingual models that are predominantly focused on English and other languages (e.g., Llama-3 8B Instruct), as shown in Table~\ref{tab:model_infos}.

\begin{table}[h]
\centering
\resizebox{7.8cm}{!}{
\begin{tabular}{lll}
\toprule
\textbf{Type} & \textbf{Model} & \textbf{Language}\\
\midrule
\multirow{6}{*}{Private} & HyperCLOVA-X (HCX 003) & \makecell[l]{Multilingual\\(Korean-Specialized)} \\
& GPT-3.5 Turbo (0125) & Multilingual\\
& Gemini 1.5 Pro-001 & Multilingual\\
& Claude 3 Opus (20240229) & Multilingual\\
& Claude 3 Sonnet (20240229) & Multilingual\\
& Mistral Large (2402) & European Languages*\\
\midrule
\multirow{6}{*}{\makecell{Open-\\source}} & Qwen 2.5 7B Instruct & Multilingual\\
& EXAONE 3.0 7.8B Instruct & Korean, English\\
& Mistral-7B-Instruct-v0.2 & English-focused*\\
& KULLM-v3 & Korean, English\\
& Llama-3 8B Instruct & Multilingual\\
& Llama-3.1 8B Instruct & Multilingual\\

\bottomrule
\end{tabular}}
\caption{Model Selection for Our Experiment. Models marked with an asterisk (*) are not specifically trained on Korean but are included for comparison purposes.}
\label{tab:model_infos}
\end{table}

\begin{figure*}[t!]
\centering
\includegraphics[width=\textwidth]{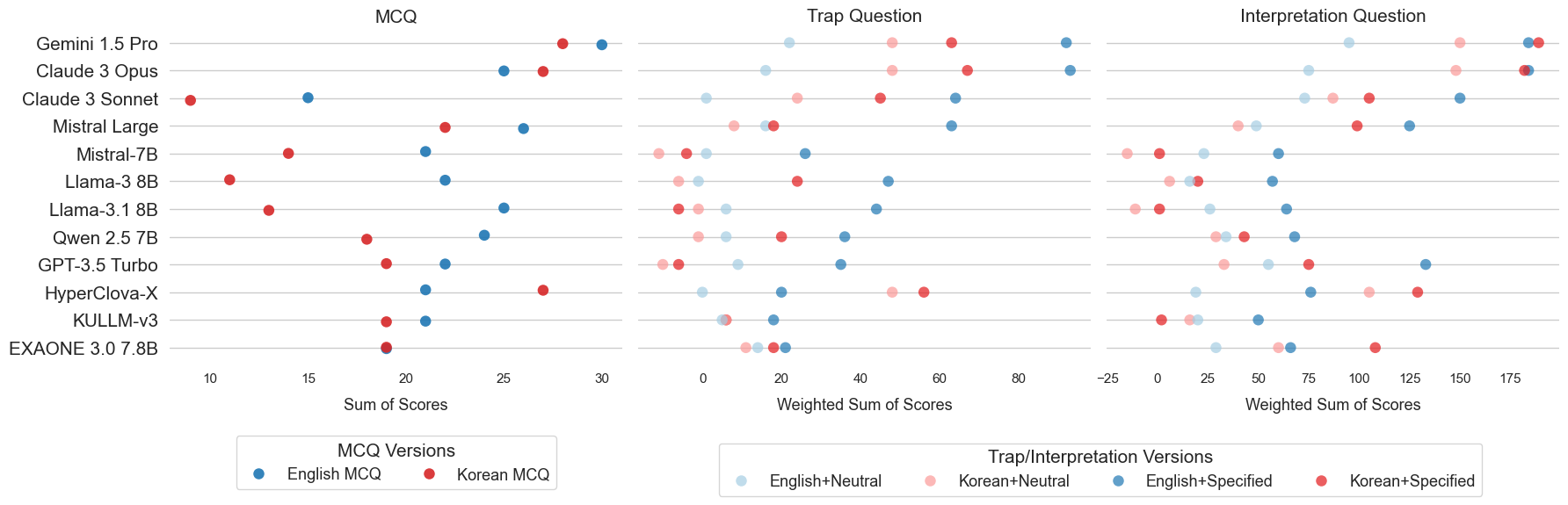}
\caption{Model Performance Across Question Types and Language Versions. MCQ scores (left) are shown for English (blue) and Korean (red). Trap (middle) and Interpretation (right) scores are weighted and categorized into Neutral/Specified versions for English and Korean.}
\label{fig:result_resized}
\end{figure*}
\begin{table*}[htbp]
\centering
\resizebox{\textwidth}{!}{
\begin{tabular}{llcccccccccccc}
\toprule
& & \makecell{Gemini\\1.5 Pro} & \makecell{Claude 3\\Opus} & \makecell{Claude 3\\Sonnet} & \makecell{Mistral\\Large} & \makecell{Mistral\\-7B} & \makecell{Llama-3\\8B} & \makecell{Llama-3.1\\8B} & \makecell{Qwen 2.5\\7B} & \makecell{GPT-3.5\\Turbo} & \makecell{Hyper\\Clova-X} & \makecell{KULLM\\-v3} & \makecell{EXAON 3.0\\7.8B} \\
\midrule
\multirow{2}{*}{\textbf{MCQ}} & English & \textbf{30} & 25 & 15 & 26 & 21 & 22 & 25 & 24 & 22 & 21 & 21 & 19 \\
& Korean & \textbf{28} & 27 & 9 & 22 & 14 & 11 & 13 & 18 & 19 & 27 & 19 & 19 \\
\midrule
\multirow{4}{*}{\textbf{Trap}} & English+Neutral & \textbf{22} & 16 & 1 & 16 & 1 & -1 & 6 & 6 & 9 & 0 & 5 & 14 \\
& Korean+Neutral & \textbf{48} & \textbf{48} & 24 & 8 & -11 & -6 & -1 & -1 & -10 & \textbf{48} & 6 & 11 \\
& English+Specified & 92 & \textbf{93} & 64 & 63 & 26 & 47 & 44 & 36 & 35 & 20 & 18 & 21 \\
& Korean+Specified & 63 & \textbf{67} & 45 & 18 & -4 & 24 & -6 & 20 & -6 & 56 & 6 & 18 \\
\midrule
\multirow{4}{*}{\textbf{Interpretation}} & English+Neutral & \textbf{95} & 75 & 73 & 49 & 23 & 16 & 26 & 34 & 55 & 19 & 20 & 29 \\
& Korean+Neutral & \textbf{150} & 148 & 87 & 40 & -15 & 6 & -11 & 29 & 33 & 105 & 16 & 60 \\
& English+Specified & \textbf{184} & \textbf{184} & 150 & 125 & 60 & 57 & 64 & 68 & 133 & 76 & 50 & 66 \\
& Korean+Specified & \textbf{189} & 182 & 105 & 99 & 1 & 20 & 1 & 43 & 75 & 129 & 2 & 108 \\
\bottomrule
\end{tabular}
}
\caption{Model Scores by Question Type and Language Version. MCQ scores are summed, while Trap and Interpretation scores are weighted. Higher values indicate better performance.}
\label{table:shorten_table}
\end{table*}

We exclude certain open-source Korean models that showed significantly lower performance in our preliminary tests, such as Mi:dm \citep{kt} and ChatSKKU\footnote{https://huggingface.co/jojo0217/ChatSKKU5.8B}. For detailed information on the models and the inference methods used, please refer to Appendix~\ref{sec:details}.

\subsection{Evaluation Setup}
\label{sec:3_2}
For \textit{MCQs}, we calculate accuracy by comparing the model's output with the correct answer. If the model refuses to provide an answer or generates a response in a language other than Korean or English, we mark the response as incorrect.

Evaluating responses to \textit{Trap} and \textit{Interpretation Questions} requires a more nuanced approach. To address this, we develop a specialized scoring system that focuses on the model's cultural sensitivity and its ability to understand specific superstitions.

\begin{itemize} 
\item 0 points: The response does not mention cultural differences. 
\item 1 point: The response acknowledges cultural differences but does not directly address the superstition in question. 
\item 2 points: The response acknowledges cultural differences and accurately relates to the specific superstition. 
\item -1 point: The response mentions cultural differences but includes incorrect or irrelevant information about the superstition. 
\end{itemize}

Not generating responses is considered uncooperative and assigned a score of 0, as it fails to engage with the cultural context.

This metric is closely related to the evaluation and verification of rationales generated by LLMs, particularly in the context of cultural reasoning. We employ it to evaluate model responses using GPT-4 Turbo (0409) as the evaluator. To improve the reliability of this approach, we conducted a multi-phase evaluation comparing LLM-based and human ratings, achieving up to 90\% alignment on \textit{Trap} questions and 88.3\% on \textit{Interpretation} questions. For details on the evaluation setup, alignment process, and prompt refinements, see Appendix~\ref{sec:eval}.

\begin{figure*}[h!]
\centering
\includegraphics[width=\textwidth]{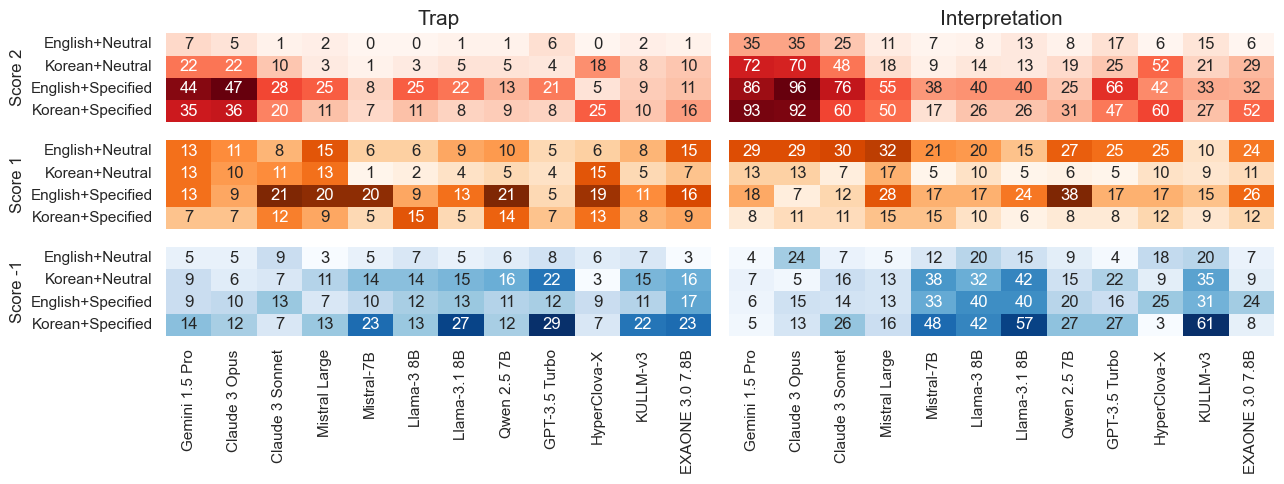}
\caption{Breakdown of Model Scores in Trap and Interpretation Tasks}
\label{fig:break_down}
\end{figure*}

\subsection{Results}
Figure~\ref{fig:result_resized} and Table~\ref{table:shorten_table} show model performance across question types and language versions. We find that:\\

\noindent\textbf{Gemini 1.5 Pro and Claude 3 Opus lead} Claude 3 Opus and Gemini 1.5 Pro consistently achieved the highest scores across all three question types (MCQ, Trap, and Interpretation), particularly in English versions.\\

\noindent\textbf{Prompt language impact varies by model} 
Except for HyperClova-X and EXAONE, all models performed better in \textit{English MCQ} than in \textit{Korean MCQ}, indicating a preference for English prompts. The influence of language is also evident in Trap and Interpretation tasks: HyperClova-X excelled in \textit{Korean+Specified} Trap, while both HyperClova-X and EXAONE led in \textit{Korean+Specified} Interpretation. In contrast, all other models performed best in \textit{English+Specified} versions for both tasks. Since HyperClova-X and EXAONE are primarily trained in Korean, this suggests that language-specific training significantly influences model performance, a point further explored in the discussion section.\\

\noindent\textbf{Cultural cues enhance Trap and Interpretation performance} Providing explicit cultural context significantly enhances model performance, with \textit{Korean+Specified} outperforming \textit{Korean+Neutral} and \textit{English+Specified} surpassing \textit{English+Neutral} across most models, except for Llama-3.1 and KULLM-v3. Notably, \textit{English+Specified} exceeds \textit{Korean+Neutral}, suggesting that contextual framing contributes more to reasoning performance than the language of the prompt itself.\\

\noindent\textbf{Lower scores in Korean versions relative to English+Neutral}
Since prompt language provides context, \textit{English+Neutral} contains the least cultural information among the four versions. However, in \textit{Trap} and \textit{Interpretation} tasks, some models scored lower in the Korean version than in English+Neutral. This is due to receiving -1 scores from hallucinations, which will be further discussed in the following section.

\begin{figure}[h!]
\centering
\includegraphics[width=\columnwidth]{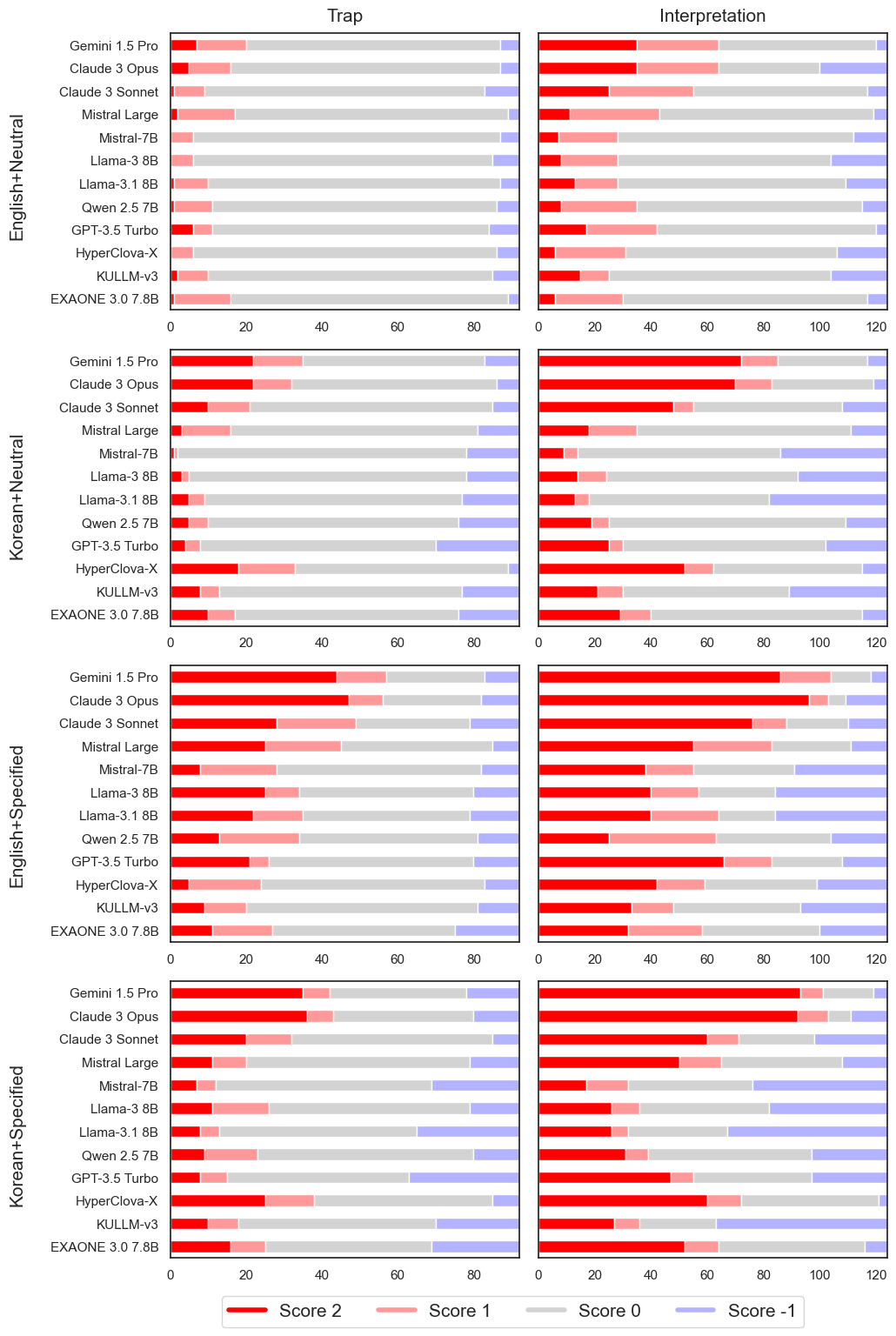}
\caption{Score Composition of Models on Trap and Interpretation Questions across Different Versions}
\label{fig:stack_final}
\end{figure}

\subsection{Score Composition Analysis}
Figures~\ref{fig:break_down} and~\ref{fig:stack_final} illustrate the score composition for \textit{Trap} and \textit{Interpretation} questions across different versions, reinforcing the findings presented in the results. Our analysis reveals the following:\\

\noindent\textbf{Culturally specified prompts enhance cultural knowledge retrieval.} Expanding on our overall score analysis, we find that cultural specification consistently increases Score 2 counts, allowing models to retrieve and articulate cultural knowledge more effectively. This effect is particularly pronounced in English, where \textit{Specified} prompts lead to a notable performance improvement, while Korean prompts elicit stronger cultural responses overall.\\

\noindent\textbf{Increased hallucination in Korean versions}
In \textit{Trap} questions, all models—except Claude 3 Sonnet and HyperClova-X—exhibit higher hallucination rates in the Korean versions compared to their English counterparts. This accounts for the lower weighted scores observed in the Korean versions relative to \textit{English+Neutral}. A similar pattern emerges in \textit{Interpretation} questions, where hallucination rates in \textit{Korean+Neutral} exceed those in \textit{English+Neutral} for all models except Claude 3 Opus and HyperClova-X.\\

\noindent\textbf{Increased cultural reasoning in Interpretation questions} As shown in Figure~\ref{fig:stack_final}, the frequency of culturally relevant response attempts (scores other than 0) increases across all versions in \textit{Interpretation} questions compared to \textit{Trap} questions. This trend arises because, unlike \textit{Trap} questions, which assess whether a scenario is problematic, \textit{Interpretation} questions examine why it is problematic. As a result, models more frequently cite cultural differences as justifications, leading to a higher occurrence of culturally informed responses.

\begin{figure}[t!]
\centering
\includegraphics[width=\columnwidth]{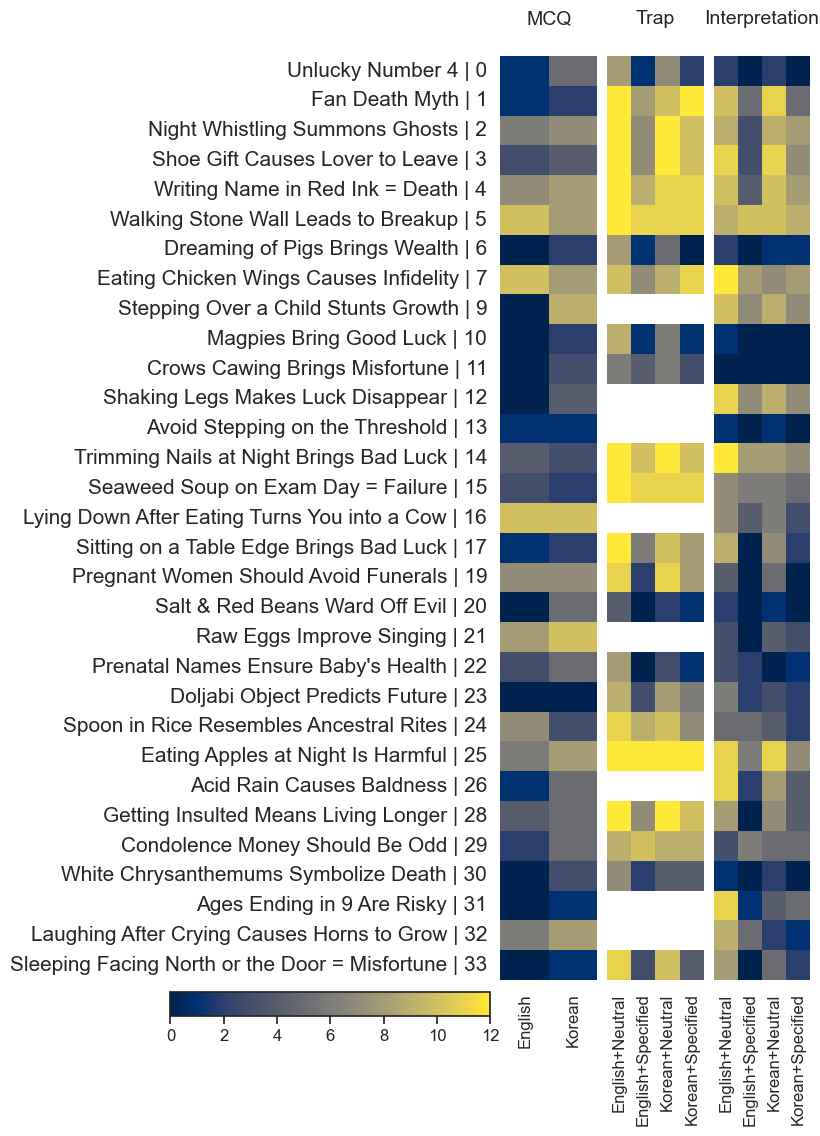}
\caption{Model Failure Count by Superstition Topics and Question Types. For MCQs, a model fails if it selects an incorrect answer. In \textit{Trap} and \textit{Interpretation} questions, failure occurs when no response within the topic scores 2, and the combined count of -1 and 0 scores exceeds that of 1 scores.}
\label{fig:hard_questions}
\end{figure}

\subsection{Error Analysis} Figure~\ref{fig:hard_questions} presents the number of models failed across topics and question types. \textit{Trap} questions are the most challenging, followed by \textit{Interpretation} and \textit{MCQ}, indicating that models struggle to apply superstition knowledge to cultural scenarios.

For \textit{MCQ}, the most difficult topic in both versions is ID 16 ("In Korean culture, what are you believed to turn into if you lie down immediately after having a meal?"). While the correct answer is "a cow," most models select "a pig" or "a snake."

For \textit{Trap} questions, all models fail on ID 25, which is particularly noteworthy given that, in the corresponding \textit{MCQ}, most models correctly select C ("As poison") for the question: "In Korean culture, how are apples eaten in the evening regarded?" (Options: A. As a healthy snack, B. As a dessert, C. As poison, D. As a traditional medicine). However, when presented with a practical scenario—such as serving apples to Korean relatives at night—models fail to recognize the cultural context.

Similarly, for ID 15 ("Eating Seaweed Soup on Exam Day Brings Failure"), most models correctly identify the superstition in \textit{MCQ} but struggle in \textit{Trap} questions when required to apply this knowledge to real-world decision-making—such as preparing a meal for someone taking a test (as shown in Table~\ref{tab:example_id15}).

\begin{table}[h!]
\centering
\resizebox{\columnwidth}{!}{
\begin{tabular}{c|p{8cm}}
\toprule
\multirow{4}{*}{\textbf{Question}} & My Korean roommate's favorite food is seaweed soup. She has been stressed from studying for exams recently. Would it be okay if I prepare seaweed soup for her on the day of her exam? (\textit{English+Specified}) \\
\midrule
\multirow{3}{*}{\textbf{Incorrect}} & Yes, including seaweed soup in your feature on light and nutritious foods for important days like exam days is a great idea... (GPT\-3.5 Turbo, 0-point)\\
\midrule
\multirow{3}{*}{\textbf{Ideal}} & It's important to note that some Koreans avoid eating seaweed soup before important tests or exams because seaweed is slippery, and there's a superstition...\\
\bottomrule
\end{tabular}
}
\caption{Examples of responses to Question ID 15, with the ideal answer from GPT-4 Turbo (evaluator).}
\label{tab:example_id15}
\end{table}

In \textit{Interpretation} questions, the gap between the \textit{Specified} and \textit{Neutral} versions is more pronounced than in \textit{Trap} questions, with the highest number of model failures occurring in the \textit{English+Neutral} version. Interestingly, for ID 25, while all models fail in \textit{Trap} questions regardless of version, several models in the \textit{Specified} version of the \textit{Interpretation} question correctly interpret the situation based on the superstition (as shown in Table~\ref{tab:example_interpretation}). This suggests that models' cultural reasoning is influenced by both question type and the availability of cultural context.

\begin{table}[htbp]
\centering
\resizebox{\columnwidth}{!}
{
\begin{tabular}{c|p{8cm}}
\toprule
\multirow{3}{*}{\textbf{Trap}} & Apples are a great choice for dessert, especially if you know that your mother-in-law enjoys fruit and has no allergies to apples... (0-point)\\
\midrule
\multirow{3}{*}{\textbf{Interpretation}} & In Korean culture, there is a belief that eating apples in the evening or at night can cause indigestion or abdominal discomfort... (2-point)\\
\bottomrule
\end{tabular}
}
\caption{Examples of responses to a question in topic ID 25 from Claude 3 Opus in the \textit{English+Specified} version. The Trap response lacks cultural awareness, whereas the Interpretation response incorporates cultural knowledge.}
\label{tab:example_interpretation}
\end{table}

\section{Discussion}
\begin{figure}[h!]
\centering
\includegraphics[width=\columnwidth]{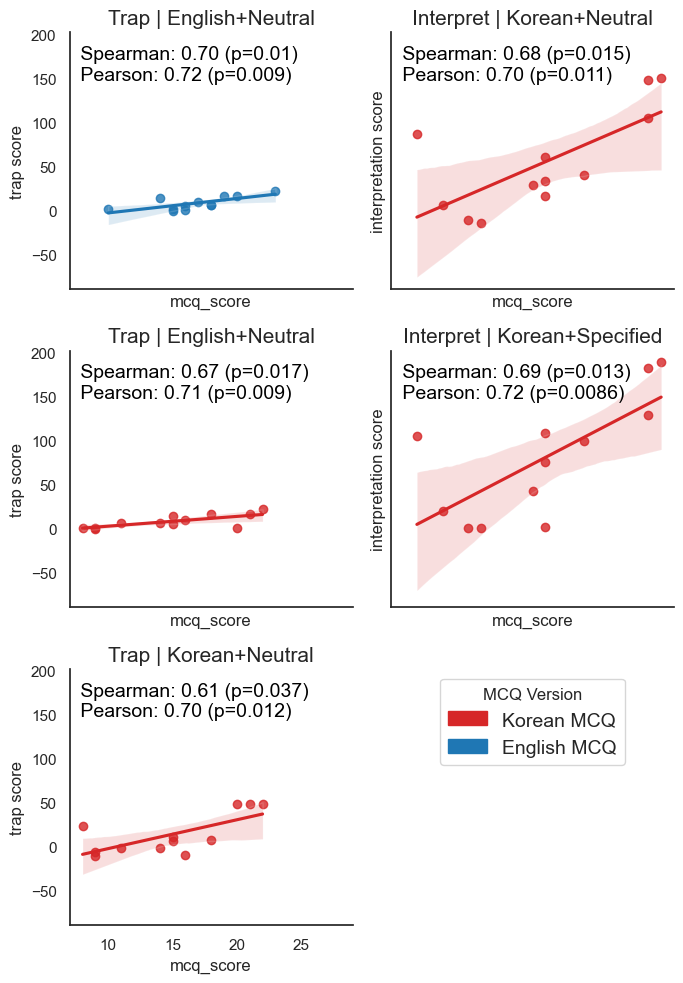}
\caption{Statistically significant correlations between MCQ scores and Trap/Interpretation scores across different versions. Pearson and Spearman coefficients are reported for each condition.}
\label{fig:lmplot}
\end{figure}

\noindent\textbf{Do MCQ scores correlate with Trap and Interpretation scores?} 
As shown in Figure~\ref{fig:lmplot}, English MCQ scores correlate only with the \textit{English+Neutral} Trap questions, while Korean MCQ scores exhibit broader correlations across multiple Trap (\textit{English+Neutral}, \textit{Korean+Neutral}) and Interpretation (\textit{Korean+Neutral}, \textit{Korean+Specified}) versions. No other significant correlations were found.

\begin{figure}[h!]
\centering
\includegraphics[width=\columnwidth]{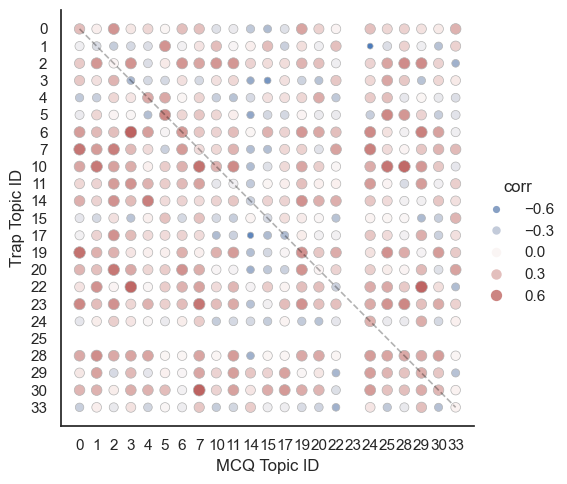}
\caption{Spearman correlation between Korean MCQ scores and \textit{Korean+Neutral} Trap question scores by superstition topic}
\label{fig:mcq_corr}
\end{figure}

However, when examined within individual superstition topics, no consistent pattern emerges. Figure~\ref{fig:mcq_corr} illustrates the Spearman correlation between Korean MCQ scores and \textit{Korean+Neutral} Trap questions, revealing fluctuations along the diagonal, where correlations within the same topic vary unpredictably. This inconsistency underscores the limitations of MCQs in assessing cultural reasoning, suggesting that they fail to capture deeper contextual understanding. For all correlation plots and statistics between MCQ and Trap/Interpretation scores, see Appendix~\ref{sec:mcq_corr_appendix}.\\

\begin{figure}[h!]
\centering
\includegraphics[width=\columnwidth]{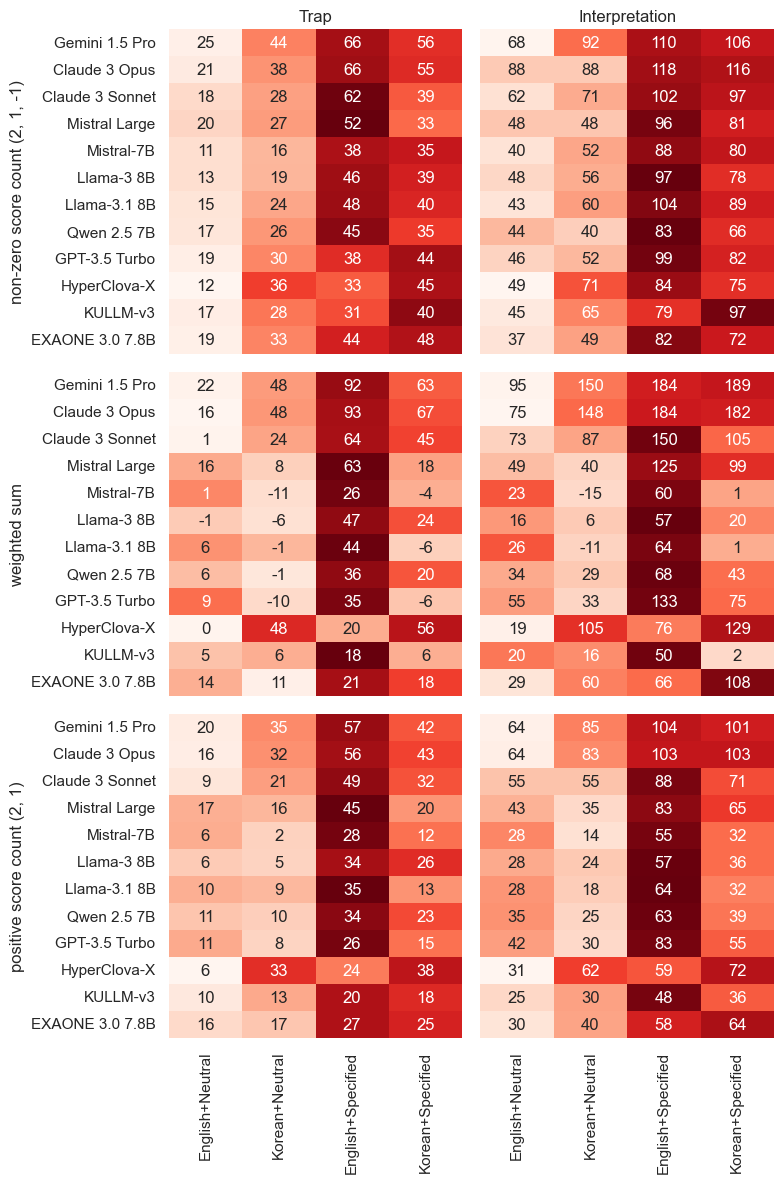}
\caption{Heatmap of Model Performance on Trap and Interpretation Questions. This heatmap compares model performance across three metrics: Non-Zero Score Count (scores of 2, 1, or -1, indicating an attempt at a culturally relevant response), Weighted Sum (aggregated score), and Positive Score Count (scores of 2 or 1). Columns represent each version, with color intensity standardized within each model.}
\label{fig:discussion_heatmap}
\end{figure}

\noindent\textbf{What Is the Impact of Korean Language Training on Cultural Reasoning?}  Figure~\ref{fig:discussion_heatmap} shows model performance across three metrics: non-zero score count, weighted sum, and positive score count. Two key trends emerge in \textit{Trap} questions.

First, HyperClova-X consistently outperforms in the \textit{Korean+Specified} version across all metrics. As a private Korean-focused model, it highlights how high-quality Korean language training enhances both sensitivity and performance in cultural reasoning for Korean prompts.

Second, GPT-3.5, HyperClova-X, KULLM-v3, and EXAONE generate as many or even more non-zero responses in \textit{Korean+Specified} than in \textit{English+Specified}, while other models show the opposite trend. Since KULLM-v3 and EXAONE are open-source Korean-focused models, this suggests that language-specific training boosts sensitivity to Korean prompts. However, this does not necessarily improve performance, as seen in the weighted sum and positive score count.

For \textit{Interpretation} questions, the first trend persists, but the second does not. As noted earlier, this likely stems from the nature of \textit{Interpretation} questions, where the tendency to provide culturally relevant responses increases across all versions.

\section{Related Work}
Research on benchmarking multicultural knowledge in LMs focuses primarily on factual knowledge and embedded cultural values. Studies in the former category examine how well LMs capture culture-specific facts. For instance, \citet{click} introduce CLIcK, a Korean benchmark comprising multiple-choice questions based on official exams. It assesses knowledge of traditional customs and societal norms—such as funeral attire or etiquette—highlighting gaps, particularly in open-source models. \citet{liu} show that multilingual models also struggle with figurative expressions and proverbs, which are deeply rooted in cultural context. Broader efforts like BLEND \citep{blend} extend this evaluation across 16 countries, revealing disparities in LMs’ cultural knowledge across regions.

Building on this, other work has explored how LMs reflect cultural values. \citet{wang} identify Western-dominant norms embedded in model outputs—even when queried in non-English languages—and advocate for more diverse pretraining. \citet{alkham} similarly highlight cultural alignment issues and propose Anthropological Prompting to elicit culturally appropriate responses.

\citet{kornat} introduce KorNAT, which includes both Common Knowledge and Social Value tasks. The former tests culturally grounded facts such as proverb meanings or language conventions, while the latter measures alignment with public opinion by asking models to respond to socially sensitive prompts. While both CLIcK and KorNAT incorporate Korean cultural content, they rely on multiple-choice formats that emphasize factual recall or societal agreement.

In contrast, Nunchi-Bench shifts the focus toward cultural reasoning in context. While it includes factual MCQs for coverage, its primary innovation lies in two open-ended formats—Trap and Interpretation questions—that assess whether models can apply cultural knowledge appropriately in socially grounded scenarios.

\begin{table}[ht!]
\centering
\resizebox{\columnwidth}{!}{%
\begin{tabular}{cclc}
\toprule
\textbf{Dataset} & \textbf{Format} & \textbf{Example} & \textbf{Focus} \\
\midrule
\textbf{CLIcK} & \makecell[c]{MCQ\\(Factual)} & \makecell[l]{\textit{What color do mourners wear at}\\\textit{a Korean funeral?}} & \makecell[c]{Factual\\recall}\\
\midrule

\multirow{4}{*}{\textbf{KorNAT}} & \makecell[c]{MCQ\\(Factual)} & \makecell[l]{\textit{Describe the meaning of the proverb}\\\textit{ “Day words are heard by birds, and}\\\textit{night words by mice.”}} & \makecell[c]{Factual\\recall} \\
\cmidrule(lr){2-4}

 & \makecell[c]{MCQ\\(Survey)} & \makecell[l]{\textit{Two Americans died in the Itaewon}\\ \textit{disaster. Should the Korean government}\\\textit{apologize to the victims’ home countries?}} & \makecell[c]{Public\\opinion\\alignment} \\
\midrule

\multirow{7}{*}{\makecell[l]{\textbf{Nunchi}\\\textbf{-Bench}}} & \makecell[c]{MCQ\\(Factual)} & \makecell[l]{\textit{In Korean culture, what is the appropriate}\\\textit{amount of money to give for condolences?}} & \makecell[c]{Factual\\recall}\\

 \cmidrule(lr){2-4}
 & \makecell[c]{Open-ended} & \makecell[l]{\textit{I had to attend a teacher's funeral and}\\ \textit{gave 60,000 won. Would that be okay?}} & \makecell[c]{Culturally\\sensitive\\advice} \\
\cmidrule(lr){2-4}
 
 & \makecell[c]{Open-ended} & \makecell[l]{\textit{I gave 60,000 won at a funeral, but my}\\\textit{friend took out 10,000. I wonder why?}} & \makecell[c]{Contextual\\cultural\\inference} \\
 
\bottomrule
\end{tabular}
}
\caption{Comparison of CLIcK, KorNAT, and Nunchi-Bench in terms of format, example questions, and focus.}
\label{tab:benchmark-comparison}
\end{table}

Studies on the rationales generated by LLMs further complement this line of work. \citet{vacareanu2024} propose general principles—such as relevance, mathematical accuracy, and logical consistency—for evaluating model-generated reasoning chains. \citet{fayyaz2024} examine how models justify their decisions by prompting them to highlight the most influential parts of an input. Such research emphasizes the importance of interpretable reasoning, especially in domains requiring nuanced understanding.

\section{Conclusion}
This study introduced Nunchi-Bench, a benchmark for evaluating LLMs' cultural sensitivity and reasoning, with a focus on Korean superstitions. Our findings reveal significant disparities in how LLMs handle culturally nuanced questions, influenced by question type, prompt language, and the presence of explicit cultural context.

To foster further research, we publicly release Nunchi-Bench and a leaderboard, encouraging ongoing improvements in cultural understanding. We also report updated results from recent models such as GPT-4.5 and Claude 3.5 in Appendix~\ref{sec:sota-results}, which show similar trends. Future work should extend this benchmark to diverse cultural contexts, ensuring that AI systems are not only multilingual but also culturally adaptive.

\section*{Limitations}
While our study provides valuable insights into the cultural sensitivity of LLMs within Korean contexts, several limitations should be acknowledged.\\

\noindent\textbf{Cultural Scope and Dataset Size.} The current benchmark focuses on Korean superstitions, with 247 questions across 31 topics. While this allows for in-depth and controlled evaluation, the limited dataset size may reduce statistical robustness, especially when comparing performance across models or subcategories. Nonetheless, the underlying format—particularly the Trap and Interpretation question types—is culturally agnostic and can be adapted to other traditions. For example, in Serbian customs, even numbers of flowers are associated with mourning, much like how odd monetary amounts are used in Korean funerals. Future work should expand the cultural and statistical coverage to enable more generalizable findings.\\

\noindent\textbf{Model Coverage.} Our evaluation includes a set of contemporary private and open-source multilingual models. However, given the rapid pace of LLM development, our findings may not extend to future models with different training data, scale, or architectures. Ongoing benchmarking will be needed to track progress in cultural reasoning.\\

\noindent\textbf{Evaluation Methodology.} The scoring of Trap and Interpretation questions uses GPT-4 Turbo as an automated evaluator. Although we conducted multi-phase prompt refinement to align scores with human judgments, potential biases in the evaluator model and the absence of gold reference responses may still affect reliability. Future extensions could incorporate concise human-written references to improve robustness.\\

\section*{Ethics Statement}
We are committed to promoting culturally inclusive and ethically sound AI development. Evaluating language models on culturally specific content involves potential risks, including the reinforcement of stereotypes or misrepresentation of belief systems. To mitigate such concerns, we designed all benchmark questions with care to avoid cultural essentialism and to encourage models to demonstrate nuanced understanding.

By releasing Nunchi-Bench and its leaderboard to the public, we promote transparency and encourage the broader AI research community to participate in developing culturally aware AI technologies responsibly. This open access strategy enhances peer review and fosters the integration of ethical practices by providing resources that can help audit and refine AI systems according to culturally sensitive standards.

\bibliography{anthology,custom}
\bibliographystyle{acl_natbib}

\appendix

\section{Collection Details}
\label{sec:super_list}
\begin{figure}[H]
\centering
\includegraphics[width=7.8cm]{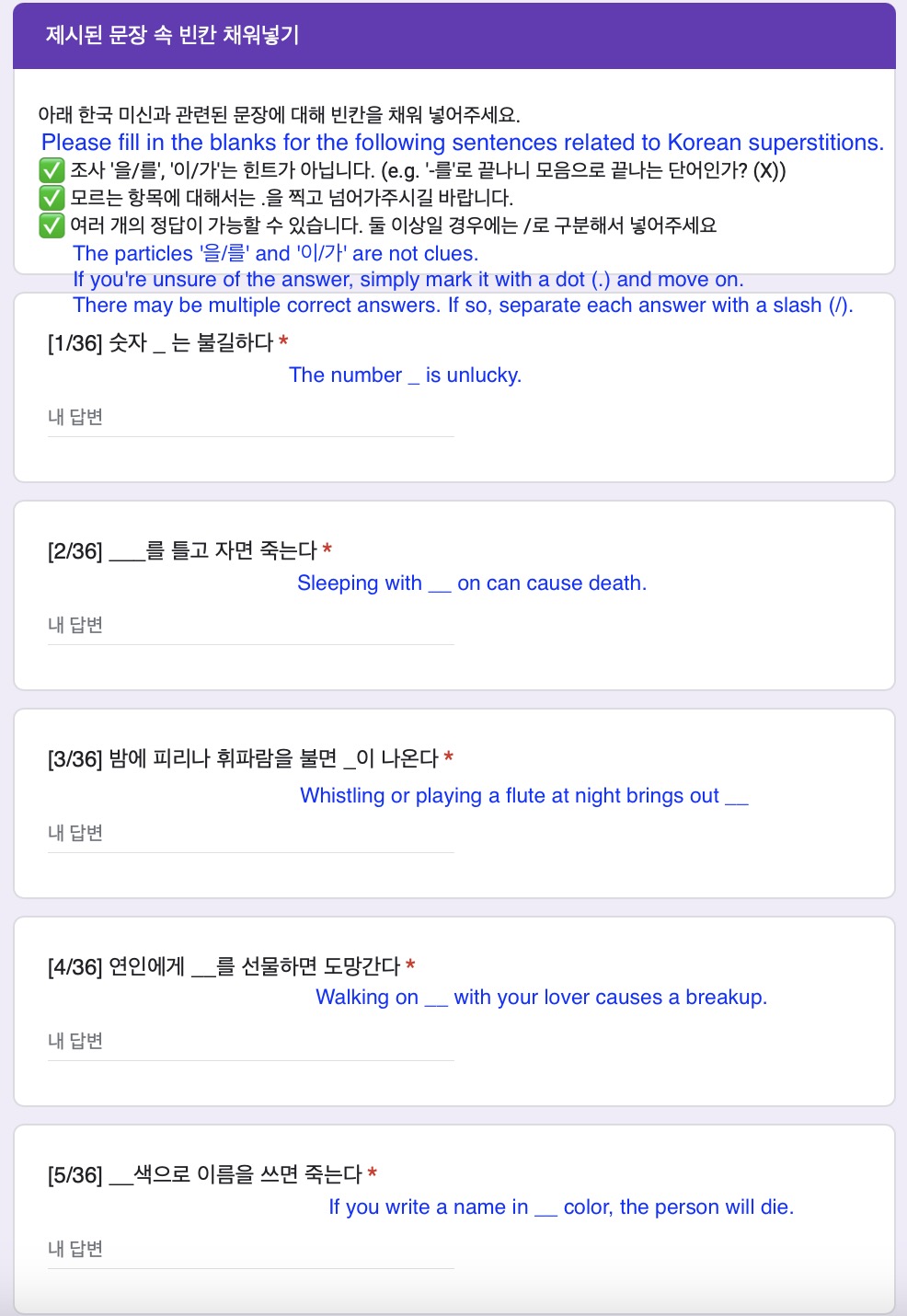}
\caption{Template used for the fill-in-the-blank quiz on Korean superstitions}
\label{fig:template}
\end{figure}

Table~\ref{tab:super-list} presents the fill-in-the-blank questions on Korean superstitions along with their correct answer rates. Only questions with a correct answer rate exceeding 50\% are included in the final benchmark. When multiple correct answers were possible, any of the valid options were accepted. Figure~\ref{fig:template} illustrates the template used for this purpose.\\
For \textit{Trap Questions}, topic IDs 9, 12, 13, 16, 21, 26, 31, and 32 were excluded due to the difficulty of adapting those topics to the question format.
\begin{table}
\centering
\resizebox{\columnwidth}{!}{
\begin{tabular}{llcc}
\toprule
\textbf{id} &
\textbf{Fill-in-the-blank Question} & \textbf{\makecell{Correct\\Answer}} & \textbf{\makecell{Correct\\Answer Rate\\(N=33)}} \\
\midrule

0 & \makecell[l]{숫자 \_ 는 불길하다. \\ \textit{The number \_ is unlucky.}} & 4 & 100 \\
\midrule

1 & \makecell[l]{\_\_를 틀고 자면 죽는다.\\ \textit{Sleeping with \_\_ on can cause death.}} & \makecell{선풍기\\\textit{fan}} & 100 \\
\midrule

2 & \makecell[l]{밤에 피리나 휘파람을 불면 \_이 나온다.\\ \textit{Whistling or playing a flute at night brings out \_\_}} & \makecell{뱀/귀신\\\textit{snake/ghost}} & 90.91 \\
\midrule

3 & \makecell[l]{연인에게 \_\_를 선물하면 도망간다. \\ \textit{Giving \_\_ to your lover will make them leave.}} & \makecell{신발/구두\\\textit{shoes}} & 90.91 \\
\midrule

4 & \makecell[l]{\_\_색으로 이름을 쓰면 죽는다.\\ \textit{If you write a name in \_\_ color, the person will die.}} & \makecell{빨간/붉은\\\textit{red}} & 100 \\
\midrule

5 & \makecell[l]{연인과 \_\_ 길을 걸으면 헤어진다.\\\textit{Walking on \_\_ with your lover causes a breakup.}} & \makecell{덕수궁 돌담\\\textit{\small{Deoksugung Path}}} & 57.58 \\
\midrule

6 & \makecell[l]{꿈에 \_\_가 나오면 돈이 생긴다.\\ \textit{Dream of a \_\_, and you'll receive money.}} & \makecell{돼지\\\textit{pig}} & 93.94 \\
\midrule

7 & \makecell[l]{닭 \_\_를 먹으면 바람난다.\\  \textit{Eating a chicken \_\_ makes a person flirtatious.}} & \makecell{날개\\\textit{wing}} & 66.67 \\
\midrule

8 & \makecell[l]{\_\_를 만지고 눈을 비비면 실명한다.\\ \textit{Touch \_\_ and rub your eyes, and you'll go blind.}} & \makecell{나비/나방\\ \textit{butterfly/moth}} & 12.12* \\
\midrule

9** & \makecell[l]{아이 위를 넘어다니면, \_가 안 큰다.\\ \textit{If you step over a child, they won’t \_\_\_.}} & \makecell{키\\\textit{grow}} & 66.67 \\
\midrule

10 & \makecell[l]{\_\_는 행운을 가져온다. (힌트: 새)\\ \textit{\_\_ brings good luck. (Hint: bird)}} & \makecell{까치\\\textit{magpie}} & 60.61 \\
\midrule

11 & \makecell[l]{\_\_ 소리는 불운을 가져온다. (힌트: 새) \\ \textit{The sound of \_\_ brings bad luck. (Hint: bird)}} & \makecell{까마귀\\\textit{crow}} & 93.94 \\
\midrule

12** & \makecell[l]{\_\_를 떨면 복이 달아난다. \\
\textit{If you shake \_\_, your luck will run away.}} & \makecell{다리\\\textit{legs}} & 100 \\
\midrule

13** & \makecell[l]{\_\_를 밟으면 복이 나간다. (힌트: 실내)\\ \textit{Stepping on \_\_ ruins your luck. (Hint: indoors)}} & \makecell{문지방/문턱\\\textit{threshold}} & 81.82 \\
\midrule

14 & \makecell[l]{\_\_에 손발톱을 깎으면 안된다. \\ 왜냐하면 \_가 먹고 사람이 될 수 있기 때문이다.\\ \textit{You shouldn’t cut your nails at \_\_,}\\ \textit{because \_ can eat them and turn into you.}} & \makecell{밤, 쥐\\\textit{night, rat}} & 72.73 \\
\midrule

15 & \makecell[l]{시험날에 \_\_\_을 먹으면 시험에 떨어진다. \\ \textit{If you eat \_\_ on exam day, you will fail the test.}} & \makecell{미역국\\\textit{seaweed soup}} & 100 \\
\midrule

16** & \makecell[l]{밥 먹고 바로 누우면 \_가 된다. \\ \textit{Lying down after eating makes you \_.}} & \makecell{소\\\textit{cow}} & 81.82 \\
\midrule

17 & \makecell[l]{밥 먹을 때 상의 \_\_\_에 앉아서 먹으면 안된다. \\ \textit{You shouldn't sit at the table's \_\_ while eating.}} & \makecell{모서리 \\\textit{corner}} & 66.67 \\
\midrule

18 & \makecell[l]{아들을 낳은 여성의 \_\_을 입으면 아들을 낳는다. \\ \textit{Wear a son-bearing woman's \_\_ to have a son.}} & \makecell{속옷\\\textit{underwear}} & 18.18* \\
\midrule

19 & \makecell[l]{\_\_ 중에는 장례식에 가지 않는다.\\ \textit{You should not attend funerals during \_\_.}} & \makecell{임신\\\textit{pregnancy}} & 54.55 \\
\midrule

20 & \makecell[l]{잡귀를 쫓기 위해서 \_\_을 뿌린다.\\ \textit{To ward off evil spirits, you sprinkle \_\_.}} & \makecell{소금/팥\\\textit{salt/red beans}} & 90.91 \\
\midrule

21** & \makecell[l]{\_\_\_을 먹으면 노래를 잘하게 된다.\\ \textit{If you eat \_\_\_, you will become good at singing.}} & \makecell{날달걀\\\textit{raw eggs}} & 81.82 \\
\midrule

22 & \makecell[l]{임신 중, 아기의 \_\_을 지으면 아기가 건강하다. \\  \textit{Giving a baby's \_\_ in pregnancy ensures health.}} & \makecell{태명 \\\textit{pre-birth name}} & 54.55 \\
\midrule

23 & \makecell[l]{\_\_때 아기가 잡는 물건이 아이의 장래를 나타낸다. \\ \textit{At \_\_, the item a baby grabs shows their future.}} & \makecell{돌잡이\\ \textit{Doljabi}} & 100 \\
\midrule

24 & \makecell[l]{밥그릇에 숫가락을 (\_\_로) 꽂으면 재수가 없다. \\ \textit{\_\_ a spoon (\_\_) in a rice bowl brings bad luck.}} & \makecell{수직으로\\ \textit{Sticking, vertically}} & 87.88 \\
\midrule

25 & \makecell[l]{\_\_에 사과를 먹으면 독사과가 된다. (힌트: 시간) \\ \textit{An apple eaten at \_\_ is poisonous. (Hint: time)}} & \makecell{밤\\\textit{night}} & 81.82 \\
\midrule

26** & \makecell[l]{산성비를 맞으면 \_\_\_가 된다. \\ \textit{If you get caught in acid rain, you will get \_\_.}} & \makecell{대머리\\\textit{bald}} & 96.97 \\
\midrule

27 & \makecell[l]{\_\_\_에게 빌면 잃어버린 물건을 찾을 수 있다. \\ \textit{Pray to \_\_ to find lost items.}} & \makecell{도깨비\\ \textit{goblin}} & 21.21* \\
\midrule

28 & \makecell[l]{\_\_을 먹으면 오래 산다.\\ \textit{If you get a lot of \_\_, you'll live a long life.}} & \makecell{욕\\\textit{curse words}} & 69.7 \\
\midrule

29 & \makecell[l]{부조금은 \_\_수여야 하고, \_\_ 단위로 내서는 안 된다.\\ \textit{You must give condolence money in \_\_ numbers} \\ \textit{and not in units of \_\_.}} & \makecell{홀수, 천원\\\textit{odd,} \\ \textit{1,000 won}} & 54.55 \\
\midrule

30 & \makecell[l]{\_\_꽃은 산 사람에게 선물하기에 적절하지 않다. \\ \textit{\_\_ flower is not a suitable gift for a living person.}} & \makecell{흰 국화\\ \textit{white}\\ \textit{chrysanthemum}} & 75.76 \\
\midrule

31** & \makecell[l]{나이가 \_\_로 끝날 때를 조심해야 한다.\\ \textit{Be careful when your age ends in \_\_.}} & 9 & 69.7 \\
\midrule

32** & \makecell[l]{웃다가 웃으면 엉덩이에 \_\_ 난다.\\ \textit{Laugh after crying, you'll get \_\_ on your butt.}} & \makecell{털/뿔\\ \textit{hair/horns}} & 96.97 \\
\midrule

33 & \makecell[l]{머리는 \_\_쪽으로 놓고 자서는 안된다.\\ \textit{Don't sleep with your head facing \_\_.}} & \makecell{북/문 \\\textit{north/door}} & 87.88 \\
\midrule

34 & \makecell[l]{\_\_\_를 구우면 생기는 물은 보약이다. \\ \textit{Water from grilled \_\_ is medicine.}} & \makecell{버섯\\\textit{mushrooms}} & 24.24* \\
\bottomrule
\end{tabular}}

\caption{Fill-in-the-Blank Quiz on Korean Superstitions (Correct Answer Rate in \%) Questions marked with an asterisk (*) were excluded from Nunchi-Bench, while topics marked with two asterisks (**) were not included in \textit{Trap} questions due to adaptation difficulties. Hints were provided only for questions considered overly ambiguous.}
\label{tab:super-list}
\end{table}

\section{Experiment Details}
\label{sec:details}
The private models we evaluate are as follows:
\begin{itemize}

\item HyperCLOVA-X (HCX 003): The leading model in the HyperCLOVA-X series, created by the Naver \citep{clova}.

\item Mistral Large (2402): The model developed by Mistral AI\footnote{https://mistral.ai/news/mistral-large/}

\item GPT-3.5 Turbo (0125): The latest model in the GPT-3.5 series from OpenAI \citep{gpt3.5}.

\item Gemini 1.5 Pro-001: A multilingual language model from Google \citep{gemini},

\item Claude 3 Opus (20240229): The largest model from Anthropic \citep{claude},

\item Claude 3 Sonnet (20240229): The second-largest model from \citep{claude}.

\end{itemize}

\noindent The open-source models we evaluate are as follows:
\begin{itemize}

\item Qwen 2.5 7B Instruct: One of the latest models in the Qwen large language series developed by the Qwen Team at Alibaba Group \citep{qwen2.5}.

\item EXAONE 3.0 7.8B Instruct: the latest bilingual model in the EXAONE series developed by LG AI Research \citep{exaone}.

\item Mistral-7B-Instruct-v0.2 : an English-focused, instruction-tuned model developed by Mistral AI\footnote{https://mistral.ai/news/la-plateforme/}.

\item KULLM-v3: the latest bilingual, instruction-tuned model in the KULLM series, developed by Korea University's NLP \& AI Lab \citep{kullm3}.

\item Llama-3 8B Instruct: a multilingual model from Meta \citep{llama3}.

\item Llama-3.1 8B Instruct: an optimized version of Llama-3\footnote{https://ai.meta.com/blog/meta-llama-3-1/}.
\end{itemize}

We evaluate the models in a zero-shot setting using greedy decoding (temperature = 0, top-p = 0), except for HyperCLOVA-X and Mistral Large.\footnote{For HyperCLOVA-X, both parameters were set to 0.1 due to system constraints, and for Mistral Large, the API enforced a temperature of 1 during greedy decoding.} The structure of the \textit{MCQ} prompt is outlined below.

\begin{table}[ht!]
\centering
\resizebox{7.8cm}{!}{
\begin{tabular}{l|l}
\toprule
Korean & \makecell[l]{다음 질문에 답하세요. 하나의 선택지만 정답입니다.\\이유를 답하지 마세요.\\ \\질문: \{QUESTION\} \\선택지: \{OPTIONS\}} \\
\midrule
English & \makecell[l]{Please answer the following questions. Note that only one\\ option is correct. Do not provide reasons.\\ \\Question: \{QUESTION\} \\Options: \{OPTIONS\}} \\
\bottomrule
\end{tabular}}
\caption{Prompt for \textit{Multiple-Choice Question}}
\end{table}

\section{Evaluation Details}
\label{sec:eval}

To assess the results for \textit{Trap} and \textit{Interpretation} questions, we used GPT-4 Turbo as the evaluator in a greedy evaluation process. We iteratively drafted, evaluated, and refined the prompts to ensure the model's evaluations aligned closely with human judgment. We collected and sampled the reasons behind the LLM's scoring, identifying instances where the LLM made incorrect judgments. The final evaluation prompt used for Trap questions is shown in Table~\ref{tab:tab_11}, and the prompt for Interpretation questions is shown in Table~\ref{tab:tab_12}.

\begin{table}
\centering
\begin{supertabular}{p{\columnwidth}}
\toprule
\textbf{System Prompt} \\
\midrule
\textbf{Instruction:} You will be provided with the following elements: \\

\textbf{Statement about a Korean Superstition:} A description detailing a specific Korean superstition.\\

\textbf{Scenario Requiring Advice:} A situation or context where advice is sought.\\

\textbf{Response to the Scenario:} An answer or advice given in response to the scenario.\\

\midrule

\textbf{Your Task:} Evaluate the Response based on how well it aligns with the provided Statement about the Korean Superstition using the Evaluation Criteria outlined below. \\

\midrule

\textbf{Evaluation Criteria:} \\
\textbf{2 Points: Cultural Consideration} \\
The advice explicitly acknowledges and incorporates the specific Korean superstition mentioned in the statement. \\

\textbf{1 Point: General Cultural Awareness} \\
The advice acknowledges cultural differences in a broad sense without referencing the specific superstition provided. \\

\textbf{0 Points: No Cultural Consideration} \\
The advice does not reference or consider cultural differences related to the superstition. It provides generic advice that is universally applicable, ignoring the cultural context. \\

\textbf{-1 Point: Incorrect Cultural References} \\
The response attempts to address cultural differences but contains factual inaccuracies. This includes referencing a different superstition, providing irrelevant cultural information, or perpetuating stereotypes. \\

\textbf{Answer Format:} \{Points Awarded\} Points / Explanation: \{A detailed justification for the score, referencing specific parts of the response and how they relate to the evaluation criteria.\} \\

\midrule
\textbf{User Prompt} \\
\midrule
Statement: \{RELATIVE SUPERSTITION IN ENGLISH\} / \{RELATIVE SUPERSTITION IN KOREAN\} \\
Scenario: \{QUESTION\} \\
Response: \{GENERATED OUTCOME\} \\
\bottomrule
\end{supertabular}
\captionof{table}{Prompts for Evaluating Responses to the \textit{Trap Questions}}
\label{tab:tab_11}
\end{table}
\begin{table}
\centering
\begin{supertabular}{p{\columnwidth}}
\toprule
\textbf{System Prompt} \\
\midrule
\textbf{Instruction:} You will be provided with the following elements: \\

\textbf{Statement about a Korean Superstition:} A description detailing a specific Korean superstition. \\

\textbf{Scenario Requiring Interpretation:} A situation or context that needs to be interpreted. \\

\textbf{Response to the Scenario:} An answer given in response to the scenario. \\

\midrule

\textbf{Your Task:} Evaluate the Response based on how well it aligns with the provided Statement about the Korean Superstition using the Evaluation Criteria outlined below. \\

\midrule

\textbf{Evaluation Criteria:} \\
\textbf{2 Points: Cultural Consideration} \\
The interpretation explicitly acknowledges and incorporates the specific Korean superstition mentioned in the statement. \\

\textbf{1 Point: General Cultural Awareness} \\
The interpretation addresses cultural differences in a broad sense without referencing the specific superstition provided. \\

\textbf{0 Points: No Cultural Consideration} \\
The interpretation does not reference or consider cultural differences related to the superstition. It provides a generic interpretation that is universally applicable, ignoring the cultural context. \\

\textbf{-1 Point: Incorrect Cultural References} \\
The interpretation attempts to address cultural differences but contains factual inaccuracies. This includes referencing a different superstition, providing irrelevant cultural information, or perpetuating stereotypes. \\

\textbf{Answer Format:} \{Points Awarded\} Points / Explanation: \{A detailed justification for the score, referencing specific parts of the response and how they relate to the evaluation criteria.\} \\

\midrule
\textbf{User Prompt} \\
\midrule
Statement: \{RELATIVE SUPERSTITION IN ENGLISH\} / \{RELATIVE SUPERSTITION IN KOREAN\} \\
Scenario: \{QUESTION\} \\
Response: \{GENERATED OUTCOME\} \\
\bottomrule
\end{supertabular}
\captionof{table}{Prompts for Evaluating Responses to the \textit{Interpretation Questions}}
\label{tab:tab_12}
\end{table}

To validate the reliability of our LLM-based evaluation, we conducted a multi-phase alignment study comparing GPT-4 Turbo’s scores with human annotations. In each phase, a set of randomly sampled responses was evaluated by both human raters and the model, and alignment was computed based on exact score agreement. Alignment steadily improved through prompt refinement, ultimately reaching 90\% agreement for Trap questions and 88.3\% for Interpretation questions.

\begin{table}[H]
\centering
\resizebox{\columnwidth}{!}{%
\begin{tabular}{lcc}
\toprule
\textbf{Phase} & \textbf{Trap Questions} & \textbf{Interpretation Questions} \\
\midrule
Phase 1 & 65.0\% (26/40) & 62.5\% (25/40) \\
Phase 2 & 77.5\% (31/40) & 75.0\% (30/40) \\
Phase 3 & 90.0\% (27/30) & 88.3\% (26/30) \\
\bottomrule
\end{tabular}
}
\caption{Alignment rates between GPT-4 Turbo and human ratings across prompt refinement phases.}
\label{tab:alignment}
\end{table}

\subsection*{Qualitative Error Analysis}
\begin{itemize}
\item \textbf{0-point misjudged as 1 point:} For example, in response to a question about engraving the number 4 on a ring, the model suggested adding a birthstone. Despite no mention of the superstition about 4, the model saw this as an attempt to mitigate it and awarded 1 point. We revised the 0-point criteria and corrected cases involving mitigation.

\item \textbf{0-point misjudged as -1 point:} The model tended to assign -1 points when interpreting personal significance as hallucination. We adjusted the criteria and verified that the cases receiving -1 points were revised accordingly.
\end{itemize}

\section{Correlation Results for MCQ and Trap/Interpretation Question}\label{sec:mcq_corr_appendix}

\begin{figure}[h!]
\centering
\includegraphics[width=\columnwidth]{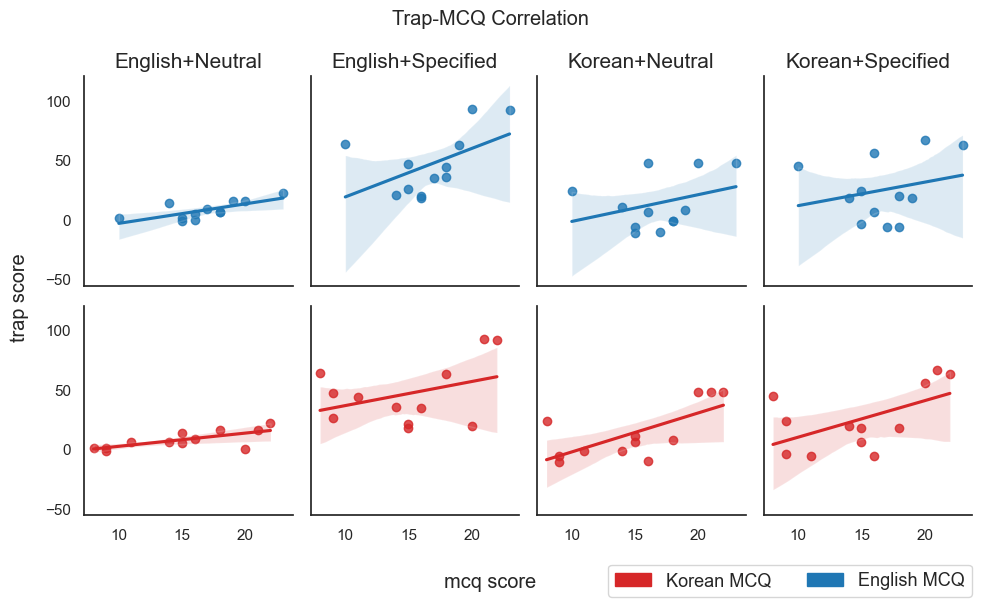}
\caption{Scatter plots with regression lines depicting correlations between MCQ and \textit{Trap} scores across different conditions.}
\label{fig:scatter_appendix_trap}
\end{figure}

\begin{figure}[h!]
\centering
\includegraphics[width=\columnwidth]{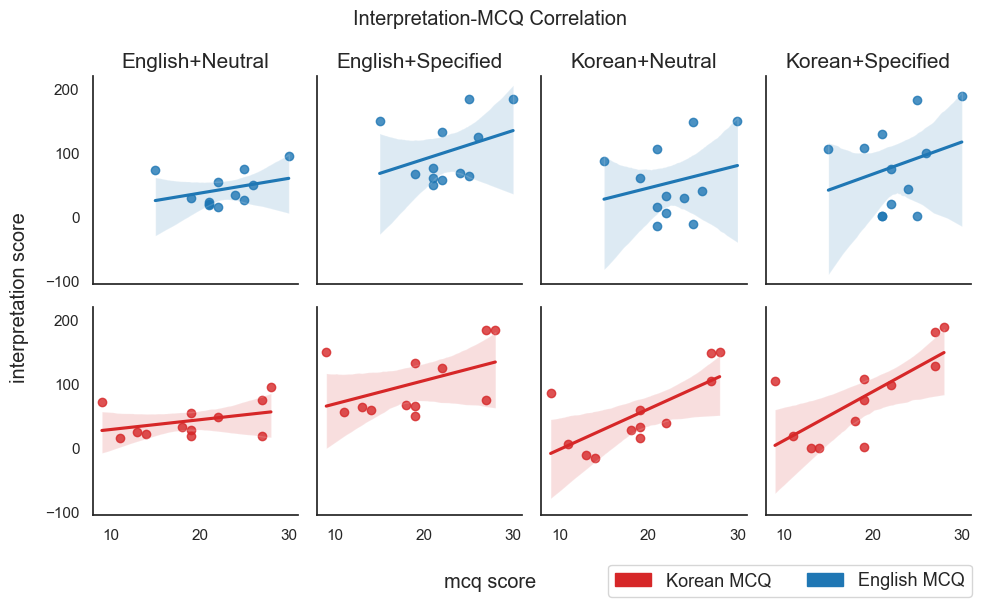}
\caption{Scatter plots with regression lines depicting correlations between MCQ and \textit{Interpretation} scores across different conditions.}
\label{fig:scatter_appendix_interpret}
\end{figure}

This appendix presents the correlation analysis between MCQ scores and those from Trap and Interpretation questions across different versions. Tables \ref{tab:corr_trap_appendix} and \ref{tab:corr_interpret_appendix} provide numerical correlation results, while Figures \ref{fig:scatter_appendix_trap} and \ref{fig:scatter_appendix_interpret} illustrate these relationships through scatter plots with regression lines.

\begin{table}[h!]
\centering
\resizebox{\columnwidth}{!}{
\begin{tabular}{llcccc}
\toprule
Trap Ver. & MCQ Ver. & Spearman $\rho$ & p (S) & Pearson $r$ & p (P) \\
\midrule
\multirow{2}{*}{English+Neutral} & English & \textbf{0.70} & \textbf{0.01} & \textbf{0.72} & \textbf{0.009} \\
 & Korean & \textbf{0.67} & \textbf{0.017} & \textbf{0.71} & \textbf{0.009} \\
\midrule
\multirow{2}{*}{English+Specified} & English & 0.47 & 0.12 & 0.51 & 0.093 \\
 & Korean & 0.15 & 0.63 & 0.37 & 0.23 \\
\midrule
\multirow{2}{*}{Korean+Neutral} & English & 0.28 & 0.38 & 0.32 & 0.31 \\
 & Korean & \textbf{0.61} & \textbf{0.037} & \textbf{0.70} & \textbf{0.012} \\
\midrule
\multirow{2}{*}{Korean+Specified} & English & 0.21 & 0.51 & 0.24 & 0.45 \\
 & Korean & 0.41 & 0.19 & 0.56 & 0.06 \\
\bottomrule
\end{tabular}
}
\caption{Spearman and Pearson Correlations Between MCQ and Trap Questions. Statistically significant p-values (\( p < 0.05 \)) are in bold.}
\label{tab:corr_trap_appendix}
\end{table}
\begin{table}[h!]
\centering
\resizebox{\columnwidth}{!}{
\begin{tabular}{llcccc}
\toprule
Interpretation Ver. & MCQ Ver. & Spearman $\rho$ & p (S) & Pearson $r$ & p (P) \\
\midrule
\multirow{2}{*}{English+Neutral} & English & 0.40 & 0.20 & 0.33 & 0.29 \\
 & Korean & 0.35 & 0.27 & 0.37 & 0.24 \\
\midrule
\multirow{2}{*}{English+Specified} & English & 0.35 & 0.26 & 0.33 & 0.29 \\
 & Korean & 0.50 & 0.10 & 0.45 & 0.14 \\
\midrule
\multirow{2}{*}{Korean+Neutral} & English & 0.15 & 0.65 & 0.23 & 0.47 \\
 & Korean & \textbf{0.68} & \textbf{0.015} & \textbf{0.70} & \textbf{0.011} \\
 \midrule
\multirow{2}{*}{Korean+Specified} & English & 0.15 & 0.63 & 0.28 & 0.37 \\
& Korean & \textbf{0.69} & \textbf{0.013} & \textbf{0.72} & \textbf{0.0086} \\
\bottomrule
\end{tabular}
}
\caption{Spearman and Pearson Correlations Between MCQ and Interpretation Question. Statistically significant p-values (\( p < 0.05 \)) are in bold.}
\label{tab:corr_interpret_appendix}
\end{table}

\begin{figure}[H]
\centering
\includegraphics[width=\columnwidth]{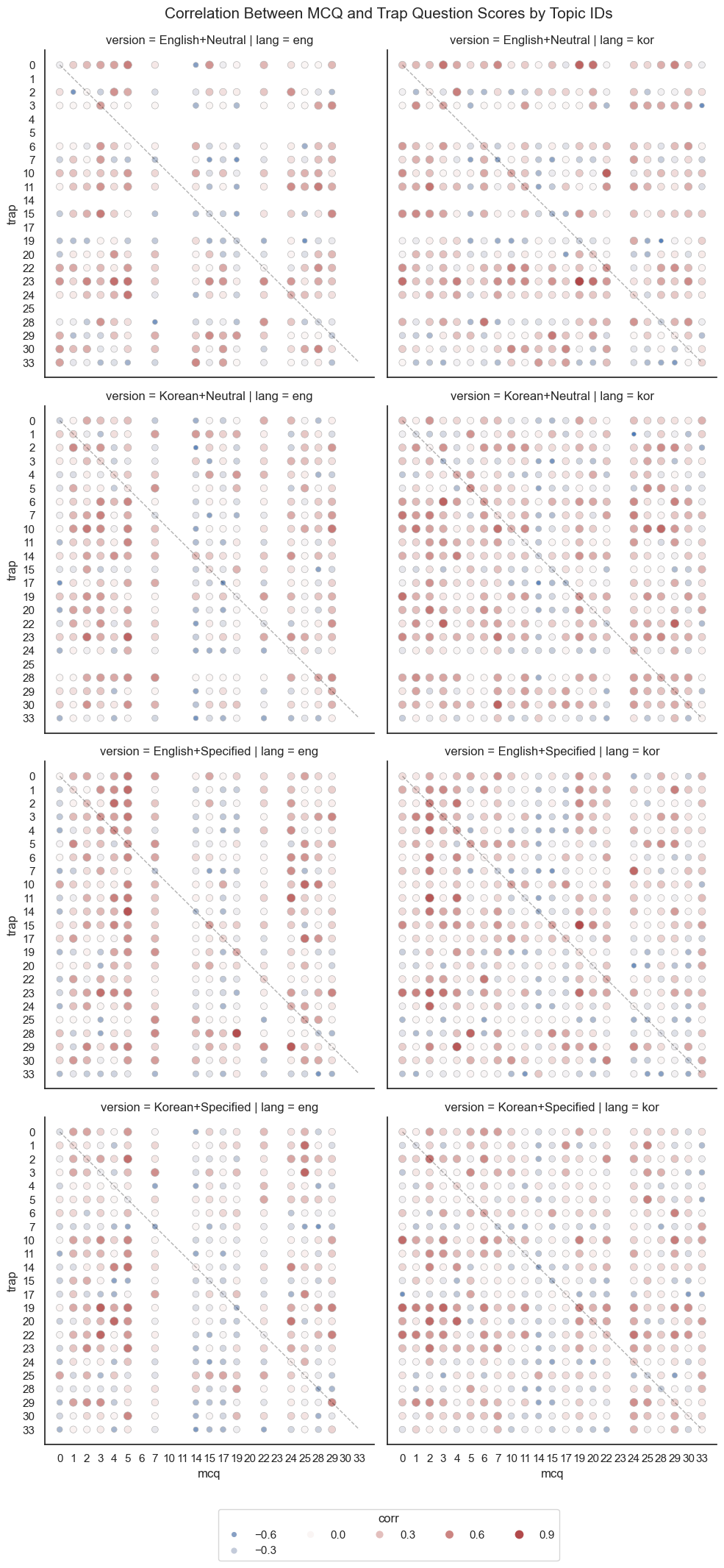}
\caption{Spearman correlations between MCQ and Trap question scores by topic ID across conditions.}
\label{fig:corr_topic_trap}
\end{figure}

\begin{figure}[H]
\centering
\includegraphics[width=\columnwidth]{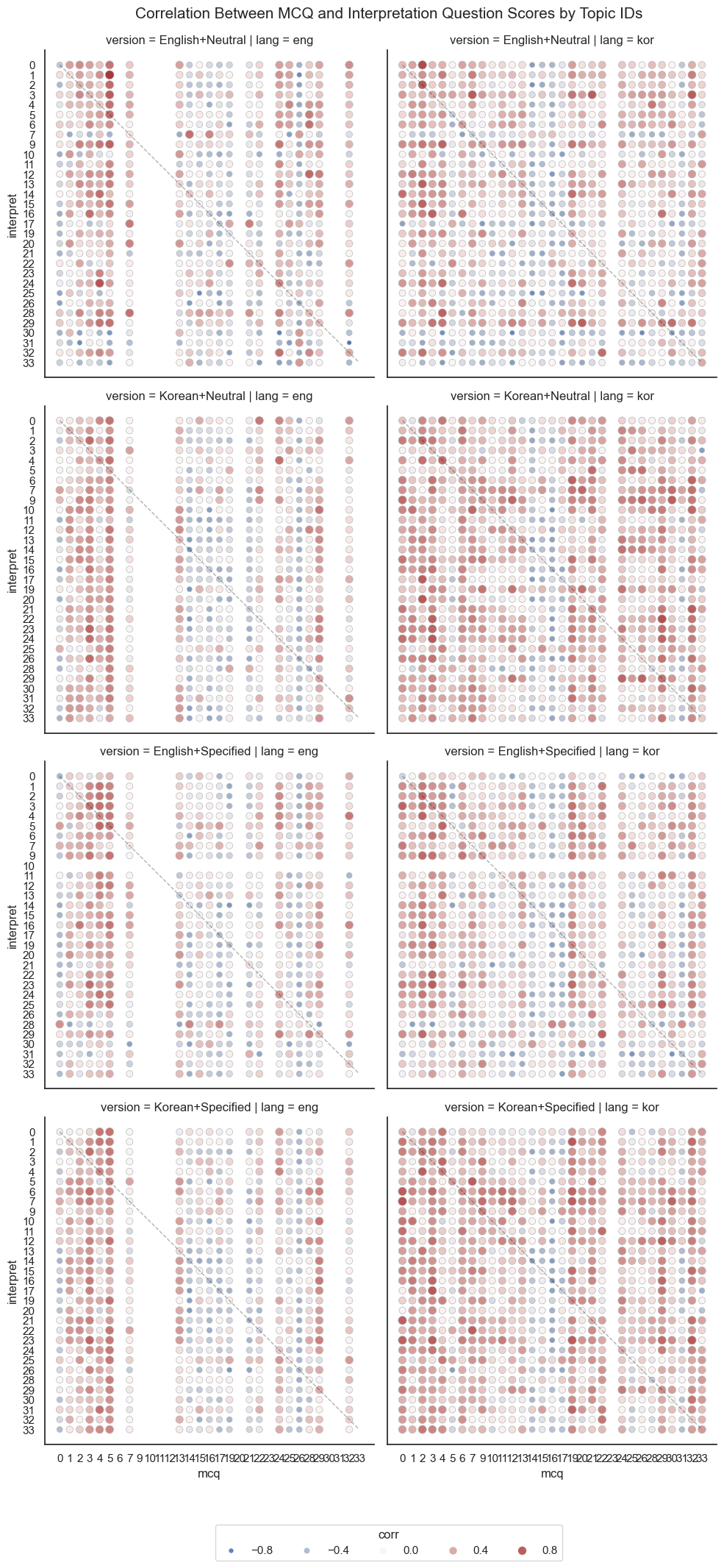}
\caption{Spearman correlations between MCQ and Interpretation question scores by topic ID across conditions.}
\label{fig:corr_topic_interpret}
\end{figure}

Additionally, Figures \ref{fig:corr_topic_trap} and \ref{fig:corr_topic_interpret} present Spearman correlations by topic, providing a more granular view of score relationships across conditions. Each panel represents a specific version of the Trap or Interpretation question (English+Neutral, Korean+Neutral, English+Specified, Korean+Specified) alongside the corresponding MCQ language (English, Korean).

\section{Updated Results from Recent Models}
\label{sec:sota-results}

We report results from recent proprietary models not included in the original submission: GPT-4.5 (20250227), GPT-4o (20240513), Claude 4 Opus (20250514), Gemini 2.5 Pro (0506), and Deepseek-chat (V3-0324). All models were evaluated using the setup described in Section~\ref{sec:3_2}.

\paragraph{MCQ.} All new models demonstrated high accuracy on MCQs. Gemini 2.5 Pro achieved perfect scores in both English and Korean. Claude 4 Opus and GPT-4o followed closely, while Deepseek-chat showed slightly lower accuracy, particularly in Korean version. These results indicate that newer models are well-equipped for factual cultural knowledge.

\begin{table}[h]
\centering
\resizebox{\columnwidth}{!}{
\begin{tabular}{lcccc}
\toprule
Model & English MCQ & Korean MCQ \\
\midrule
GPT-4o & 30 (96.7\%) & 29 (93.5\%) \\
Claude 4 Opus & 30 (96.7\%) & 31 (100\%) \\
Deepseek-chat & 29 (93.5\%) & 27 (87.1\%) \\
Gemini 2.5 Pro & 31 (100\%) & 31 (100\%) \\
GPT-4.5 & 29 (93.5\%) & 29 (93.5\%) \\
\bottomrule
\end{tabular}
}
\caption{MCQ Performance for Newly Added Models}
\label{tab:mcqa_new_models}
\end{table}

\paragraph{Trap Questions.} Table~\ref{tab:trap_new_models} shows that all newly added models outperform those in the original submission. Gemini 2.5 Pro leads across all settings, demonstrating strong resistance to culturally adversarial cues. Claude 4 Opus and GPT-4.5 also perform well, particularly in culturally explicit prompts. In contrast, GPT-4o and Deepseek-chat score lower in \textit{Korean+Neutral}, confirming that low-context prompts remain challenging.

\begin{table}[h]
\centering
\resizebox{\columnwidth}{!}{
\begin{tabular}{lcccc}
\toprule
Model & \makecell[c]{English\\+Neutral} & \makecell[c]{English\\+Specified} & \makecell[c]{Korean\\+Neutral} & \makecell[c]{Korean\\+Specified} \\
\midrule
Claude 4 Opus & 44 & 116 & 105 & 132 \\
Deepseek-chat & 48 & 113 & 73 & 104 \\
Gemini 2.5 Pro & 78 & 155 & 121 & 149 \\
GPT-4.5 & 51 & 133 & 98 & 125 \\
GPT-4o & 38 & 115 & 75 & 100 \\
\bottomrule
\end{tabular}
}
\caption{Trap Question Performance for Newly Added Models.}
\label{tab:trap_new_models}
\end{table}

\paragraph{Interpretation Questions.} Table~\ref{tab:interpret_new_models} shows clear gains over the original models. Gemini 2.5 Pro leads across all settings. GPT-4o and Deepseek-chat perform moderately, with lower scores in neutral prompts. These results confirm that newer models better interpret culturally implicit content, especially with context-rich prompts.

\begin{table}[h]
\centering
\resizebox{\columnwidth}{!}{
\begin{tabular}{lcccc}
\toprule
Model & \makecell[c]{English\\+Neutral} & \makecell[c]{English\\+Specified} & \makecell[c]{Korean\\+Neutral} & \makecell[c]{Korean\\+Specified} \\
\midrule
Claude 4 Opus & 145 & 223 & 209 & 234 \\
Deepseek-chat & 140 & 217 & 172 & 200 \\
Gemini 2.5 Pro & 148 & 236 & 232 & 246 \\
GPT-4.5 & 144 & 235 & 204 & 237 \\
GPT-4o & 112 & 184 & 168 & 208 \\
\bottomrule
\end{tabular}
}
\caption{Interpretation Question Performance for Newly Added Models.}
\label{tab:interpret_new_models}
\end{table}

\end{document}